%% file: RSS_Workshop.tex
\documentclass[conference]{IEEEtran}
\usepackage{times}

\usepackage[numbers]{natbib}
\usepackage{multicol}
\usepackage[bookmarks=true]{hyperref}
\usepackage{gensymb}
\usepackage{enumitem}
\usepackage{amsmath}
\usepackage{amssymb}
\usepackage{hyperref}

\hypersetup{
    colorlinks=true,
    linkcolor=blue,
    filecolor=magenta,      
    urlcolor=blue,
    pdftitle={Overleaf Example},
    pdfpagemode=FullScreen,
    }

\usepackage{graphicx}
\usepackage{float} 
\usepackage{subfigure}
\usepackage{dirtytalk}
\usepackage{multirow}
\usepackage{booktabs}
\usepackage{array}

\usepackage{comment}
\usepackage{ulem}

\usepackage{wrapfig}
\usepackage{tikz}
\usetikzlibrary{calc}

\usepackage{etoolbox}
\makeatletter
\patchcmd{\@makecaption}
  {\scshape}
  {}
  {}
  {}
\makeatother

\begin{document}

\title{Evolutionary Curriculum Training for DRL-Based Navigation Systems}
\author{
  Max Asselmeier$^{*}$, Zhaoyi Li$^{*}$, Kelin Yu$^{*}$, Danfei Xu \\
  Institute for Robotics and Intelligent Machines \\
  Georgia Institute of Technology, Atlanta, Georgia 30332–0250 \\
  Email: \{mass, zhaoyi, coliny, danfei\}@gatech.edu \\
  $^{*}$ Equally Contributed, Alphabetical Order
  }



%

\maketitle
\vspace{-0.5cm}
\begin{abstract}
In recent years, Deep Reinforcement Learning (DRL) has emerged as a promising method for robot collision avoidance. However, such DRL models often come with limitations, such as adapting effectively to structured environments containing various pedestrians. In order to solve this difficulty, previous research has attempted a few approaches, including training an end-to-end solution by integrating a waypoint planner with DRL and developing a multimodal solution to mitigate the drawbacks of the DRL model. However, these approaches have encountered several issues, including slow training times, scalability challenges, and poor coordination among different models. To address these challenges, this paper introduces a novel approach called evolutionary curriculum training to tackle these challenges. The primary goal of evolutionary curriculum training is to evaluate the collision avoidance model's competency in various scenarios and create curricula to enhance its insufficient skills. The paper introduces an innovative evaluation technique to assess the DRL model's performance in navigating structured maps and avoiding dynamic obstacles. Additionally, an evolutionary training environment generates all the curriculum to improve the DRL model's inadequate skills tested in the previous evaluation. We benchmark the performance of our model across five structured environments to validate the hypothesis that this evolutionary training environment leads to a higher success rate and a lower average number of collisions. Further details and results at our project website\footnote[1]{\href{https://sites.google.com/view/hierarchical-navigation}{https://sites.google.com/view/hierarchical-navigation}}.
\end{abstract}

\IEEEpeerreviewmaketitle

\input{TextFiles/Introduction}
\input{TextFiles/RelatedWorks}

\input{TextFiles/ProblemFormulation}
\input{TextFiles/Methodology}

\input{TextFiles/Experiments}
\input{TextFiles/Conclusion}

\bibliographystyle{plainnat}
\bibliography{references}

\input{TextFiles/Appendix}

\end{document}

%% file: TextFiles/Introduction.tex
\section{\textbf{Introduction}}
Decentralized Multi-Agent Collision Avoidance is a critical aspect of robot motion planning and has long been an unsolved problem. In recent years, Deep Reinforcement Learning (RL) has shown great success in multi-agent coordination and collision avoidance applications. Previous works on Multi-Agent Reinforcement Learning (MARL) collision problems have primarily focused on utilizing a learned approach to act upon local sensor or agent data and generate control actions \cite{Everett_Motion, Everett_Collision, long_towards_2018, Semnani_Multi_agent}. However, such pure DRL approaches lack the ability to conduct long-horizon planning effectively.  In order to address the aforementioned limitations, advanced approaches integrate a higher-level planner for the identification of waypoints and global goals while simultaneously employing a lower-level DRL model to optimize local decision-making \cite{Brito_Where, Krantz_Waypoint, kastner2021arena, 9811797}. This hierarchical approach significantly enhances the DRL model's ability to perform long-horizon planning. 

Yet, despite the progress made in this area, the DRL-based collision avoidance models still possess certain constraints. While their hierarchical structures allow for effective long-horizon planning, the lower-level DRL models encounter limitations when it comes to executing sharp turns or U-turns in structured environments and maintaining collision avoidance consistently over extended periods. These factors are crucial for the practical implementation of robots in real-world scenarios. Recent approaches have addressed this problem through two main strategies. The first approach involves incorporating a high-level planner into the training pipeline of the Deep Reinforcement Learning (DRL) model \cite{kästner2023holistic}. Another approach trains a model to switch between classical model-based navigation and DRL model based on the environmental scenario \cite{9811797}. However, these approaches suffer from the challenges of being either excessively difficult to train or encountering synchronization issues between the different behaviors. These methods primarily focus on circumventing the need for enhancing the existing DRL model, instead relying on an alternative model to handle tasks that the DRL model may not excel at.

On that account, we introduce an evolutionary curriculum training environment that aims to optimize the capabilities of lower-level DRL models, ensuring their adaptability across multiple scenarios. Our contribution comprises two essential components. Firstly, we have developed an evaluation process to comprehensively assess the DRL model's performance and limitations across diverse environments. Secondly, we have established a curriculum-based training environment for the DRL model that specifically targets and enhances any identified deficiencies.

In previous research, the simulation environments used by DRL models failed to accurately replicate the level of complexities found in real-world environments, which hinders the generalizability and scalability of the corresponding DRL models. To address these limitations and improve the evaluation and training of DRL models, we have designed and implemented a random environment generator that closely emulates real-world scenarios and aims to enhance two main capabilities of the agent. First, its ability to navigate through structured maps, adeptly maneuvering through rooms of varying shapes while effectively avoiding static obstacles. Second, its ability to avoid dynamic obstacles such as pedestrians or other robots. To augment these two environment features, we have implemented multiple design variables that allow us to control the generation of the environment:

\begin{itemize}
\item 
\textbf{Map structure}: A room/maze generator that constructs multiple rooms connected by corridors. Map generation is controlled by the parameters room count, room size, corridor width, and convexity level.
\item \textbf{Dynamic Obstacles}:  A pedestrian generator that instantiates diverse pedestrians within the environment. The generation of these pedestrians is governed by the parameters pedestrian count, pedestrian speed, and pedestrian policy.
\end{itemize}
Utilizing this simulation environment, we assess the change in performance of the DRL model ($\Delta$ \textbf{PerfScore}) in response to changes in environment variables. Through analyzing the $\Delta$ \textbf{PerfScore}, we identify the types of environments in which the DRL model encounters challenges. The evolutionary training environment uses the corresponding environment variables to generate curricula specifically tailored to address those environments. The generated curricula are designed to progressively increase in difficulty, aiming to accelerate training and mitigate the risk of overfitting. 

In addition to employing evolutionary curriculum training, we have integrated an local waypoint planner into the Evolution-DRL model that was trained using our evolutionary curriculum training approach. The waypoint planner provides a coarse navigation plan without considering other agents or collision avoidance, while the Evolution-DRL model offers short-distance navigation with awareness of dynamic agents. This combination enables the Evolution-DRL model to possess long-horizon planning abilities which we verify through long-range navigation tests. Through a comprehensive experimental analysis, we have observed that our evolutionary curriculum training method not only enhances the performance of the DRL model itself but also significantly improves the performance of the hierarchical navigation model, elevating it to a state-of-the-art level \cite{9811797, kästner2023holistic}. \\ 
\vspace{-0.3cm}

%% file: TextFiles/RelatedWorks.tex
\section{\textbf{Related Works}} \label{sec:related_works}

\subsection{\textbf{Multi-agent Collision Avoidance}}
The task of multi-agent collision avoidance is one of paradigm importance in robot motion planning. Because of this, it has received a great deal of attention. Many works have sought to tackle the task through classical planning approaches \cite{4621214, fiorini1998motion, maren_DWA, vandenburg_recipriocal_2011}, but given that the proposed work revolves around DRL, we will mostly focus on learning-based approaches. 

Within the learning-based approaches to multi-agent collision avoidance, a separation can be made between agent-level and sensor-level architectures, where agent-level models employ agent state information as the input to the avoidance policy \cite{Brito_Where, Everett_Collision, Semnani_Multi_agent, chen_crowd_robot_2018, chen_socially_2017} and sensor-level models employ sensor data as the policy input \cite{long_towards_2018, tai_socially_2018, chiang_learning_2018, kastner2021arena}. Agent-level approaches elude the need to reason about static and dynamic obstacles by solely reasoning about other agents but subsequently limit their collision avoidance capabilities to avoiding other agents. This means that such a policy must be used as a single component within a larger framework of long-range navigation is to be performed. Sensor-level approaches are able to avoid all obstacles in the environment but at the cost of feeding in high-dimensional scans and images as the model input. Some newer approaches attempt to deploy several collision avoidance strategies that are engaged depending on environmental circumstances \cite{9811797, chiang_learning_2018}, but such an approach has limited scalability due to the need to implement each strategy and identify when it should be deployed.

\vspace{-0.2cm}

\subsection{\textbf{Curriculum Learning}}
Curriculum learning \cite{ELMAN199371, curriculum} revolves around training a model using progressively more difficult data, reflecting the organization of academic curricula consumed by humans. With the use of curriculum learning comes the need to a.) gauge the “difficulty” of training data and b.) schedule the escalation of the data difficulty during training so that the model is given challenging scenarios to sophisticate its performance while not making such scenarios overly complex and preventing the agent from learning anything. 

This approach has seen use in object detection and classification \cite{9008373}, natural language processing \cite{curr_nlp}, and robotics \cite{curr_value}. Some robotics applications do involve adjusting the training environment, but for tasks such as locomotion \cite{wang_poet_2019}. Some approaches have been used for navigation and collision avoidance, but in regards to adjusting the initial robot state \cite{ivanovic_barc_2018} or the number of agents \cite{long_evolutionary_2020}. In this proposed work, we investigate how the evolution of the training environment itself impacts multi-agent collision avoidance performance.

%% file: TextFiles/Methodology.tex
\section{\textbf{Methodology}} \label{sec:methodology}
Firstly,  in Section \ref{sec:environment_setup}, we detail how we designed and created simulation environments for training. Secondly, in Section \ref{sec:perf_score}, we introduce our approach for designing the performance score \textit{PerfScore} used in the evaluation and training process. In section \ref{sec:evaluation_method}, we cover how we evaluate agent performance for our curriculum learning approach, and in Section \ref{sec:evolutionary_training}, we detail our evolutionary training pipeline. A diagram of the overall planning pipeline is shown in Figure \ref{fig:diagram}. Some preliminary knowledge can be founded in Appendix \ref{sec:problem_formulation}. Technical and implementation details for the waypoint planner can be found in Appendix \ref{sec:local_waypoint_planner}

\subsection{\textbf{Training Environment Generation}} \label{sec:environment_setup}


We have divided the design variables of the training environment generator into two primary parts, map structure and dynamic obstacles.

\begin{figure}[ht]
    \centering
    \includegraphics[width=0.5\textwidth]{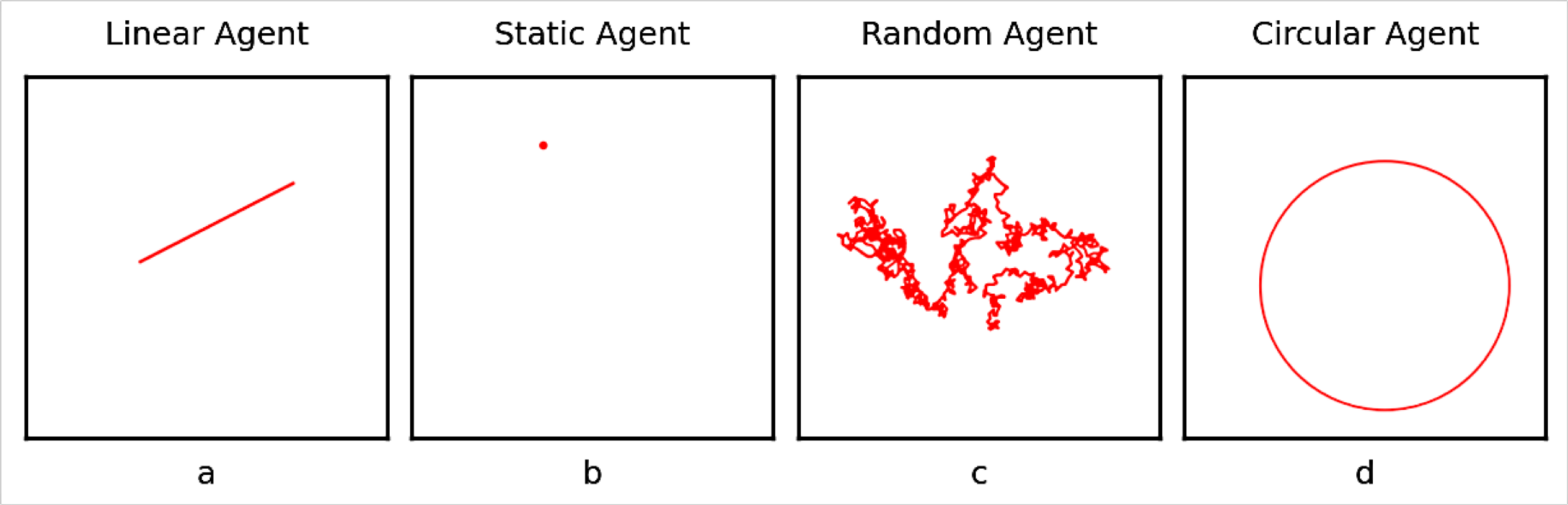}
    \vspace{-0.5cm}
    \caption{
            Sample trajectories of different pedestrian policies.}
    \label{fig:traj}
\end{figure}

\begin{itemize}
\item 
\textbf{Map Structure}: To instantiate the map structure, we have developed a room and maze generator capable of producing interconnected rooms linked by corridors. The map generator incorporates the following four design variables to enable the diverse generation of maps (examples in Figure \ref{fig:comparasion}):

\begin{itemize}
\item \textbf{Room Number} Room number indicates the number of individual rooms generated in the environment. For DRL model evaluation and training, we limit the range to $[0, 4]$ due to the lack of long-horizon planning skills.

\item \textbf{Room Size} The room size variable is a scalar value on the range $[0, 1]$ that is able to modify both the average and maximum room sizes. For model training and evaluation, we limit the range to [0.5, 1], as to not generate rooms that are excessively and unrealistically small.

\item \textbf{Corridor Width} Similar to room size, corridor width is a scalar value on the range $[0, 1]$ that can adjust all the generated corridors' widths. For model evaluation and training, we limit the range to $[0.5, 1]$, as it is unrealistic for a corridor to be excessively narrow.

\item \textbf{Convexity} Convexity is a positive scalar value that indicates how rooms are connected. A lower convexity value indicates a reduced likelihood of sharp turns in the map. If convexity is 1, the connections appear as a zigzag between rooms, and if convexity is $\infty$, the connection between two rooms must be either straight or paths that are perpendicular.  For model training and evaluation, we restrict the selection of convexity levels to $[1, 2, 3, 4, \infty]$.

\end{itemize}

\begin{figure}[ht]
    \centering
    \includegraphics[width=0.45\textwidth]{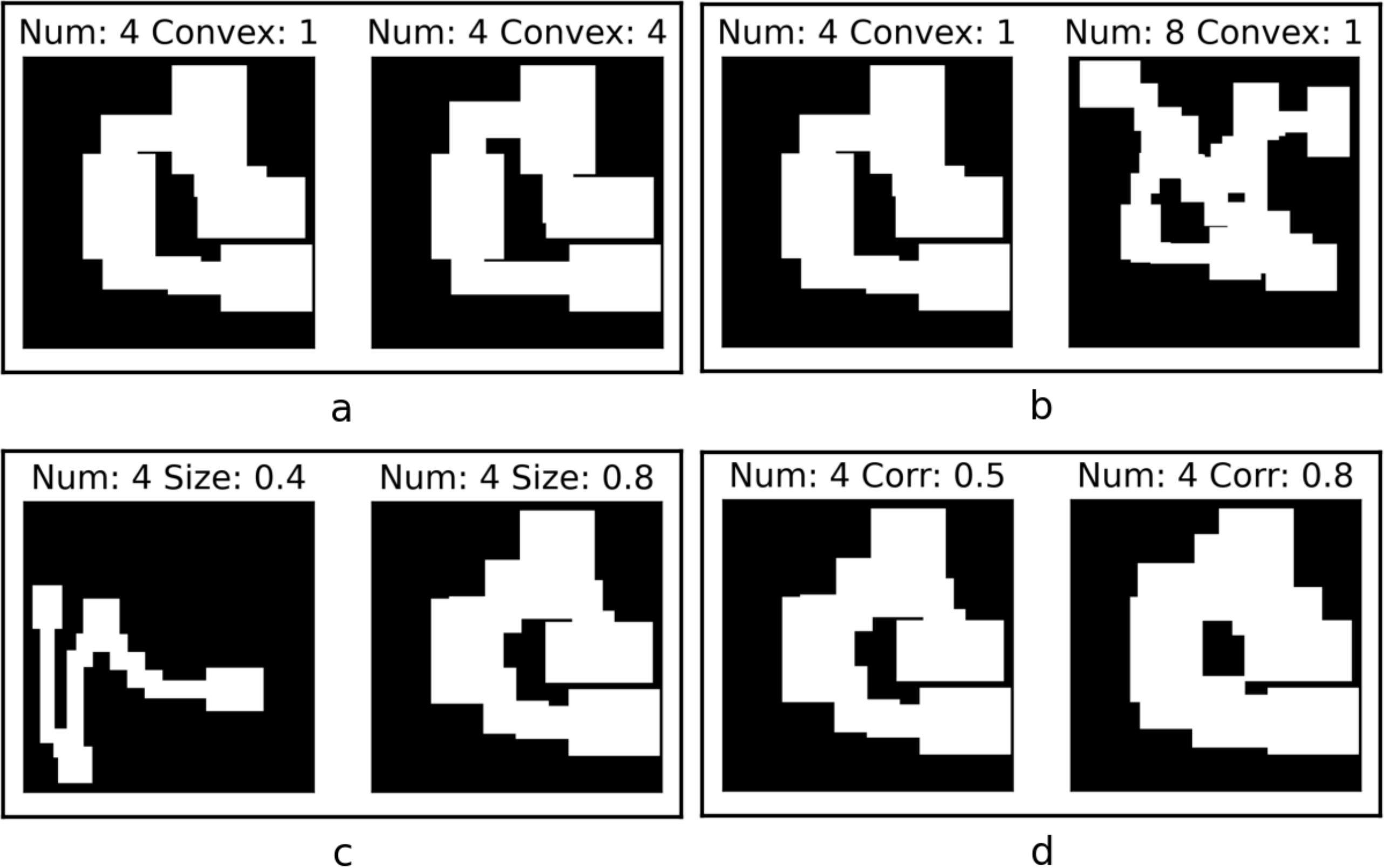}
    \caption{
            Comparisons between maps with different variables. Subfig. a is a comparison between different Convexity. Subfig. b is a comparison between different Room Number. Subfig. c is a comparison between different Room Size. Subfig. d is a comparison between different Corridor Width.}
    \label{fig:comparasion}
\end{figure}

\item 
\textbf{Dynamic Obstacles}: To generate the dynamic obstacles, we have developed a pedestrian generator that spawns a diverse set of agents in the environment. Examples of such agents are shown in Figure \ref{fig:traj}. The pedestrian generator incorporates the following three design variables:

\begin{itemize}
\item \textbf{Pedestrian Number} Pedestrian numbers indicate the number of pedestrians that will be generated in the environment. For model training and evaluation, we limit the range to $[10, 18]$ in order to avoid an excessively crowded environment.

\item \textbf{Pedestrian Speed} In simpler pedestrian policies, the pedestrian speed remains constant, resulting in all pedestrians with such policies having the same speed. However, in more complex policies such as circle walking and random walking, the pedestrian speed influences the average value of the randomly generated speeds. For model training and evaluation, we restrict the range to $[1, 2]$ since it is unrealistic for pedestrians to move at extremely high speeds.

\item \textbf{Pedestrian Policy}  The pedestrian policy variable, ranging from $[0.0, 0.8]$, indicates the percentage of challenging policies, including circle walking and random walking, present in the environment. Higher values of pedestrian policy variables correspond to a greater proportion of challenging policies.

\end{itemize}
\end{itemize}

\subsection{\textbf{Performance Score Formulation}} \label{sec:perf_score}

To gauge training environment difficulty, we define a unified \textit{PerfScore} as both the evaluation metric and reward function for learned collision avoidance model.

\begin{equation}
    P(\mathbf{s_t}) = \begin{cases}
            1 & \text{if $\mathbf{r_{pos}} = \mathbf{r_{goal}}$}\\
            -0.25 & \text{if $g_{min} < 0$ and $o_{speed} = 0$} \\
            -(1-e^{d_{\text{min}}-0.3}) & \text{if $g_{min} \leq 0.3$ and $o_{speed} \neq 0$} \\
            0 & \text{otherwise}
    \end{cases}
\end{equation}

The \textit{PerfScore} consists of several score equations specifically designed to evaluate certain skills of agents and encourage improvement on those skills during training. The \textit{PerfScore} is assigned to 1 when the agent successfully reaches the goal ($\mathbf{r_{pos}} = \mathbf{r_{goal}}$). This equation evaluates the agents' ability to perform short-distance navigation while providing incentives for them to learn and improve navigation skills during training. The \textit{PerfScore} is set to -0.25 when agents collide ($g_{min} < 0$) with any static obstacles such as walls, where $o_{speed}$ represents the obstacle velocity ($o_{speed} = 0$). This equation evaluates the agents' ability to avoid static obstacles,  while also offering incentives for them to learn and improve their obstacle avoidance skills during training. Similarly, we evaluate the agents' ability to avoid dynamic obstacles ($o_{speed} \neq 0$), such as pedestrians. However, with static obstacles, the reward issued initiates a decrease in the score before any actual collision occurs ($g_{min} \leq 0.3$). This adjustment is made because it is essential to maintain a safe distance from pedestrians at all times. To promote social awareness in our agents, the \textit{PerfScore} is intentionally lower ($-(1-e^{d_{\text{min}}-0.3})$) when collisions occur with pedestrians compared to collisions with walls. If none of these occur at timestep $t$, then no reward is given.

\subsection{\textbf{Performance Evaluation}} \label{sec:evaluation_method}

To examine the influence of various environmental variables on model's \textit{PerfScore}, we employ a simple sampling method. For each environment variable, we pick the minimum and maximum from its range, and for each sampled minimum and maximum, we generate 500 maps corresponding to the specific configuration, simulate 5000 iterations, and collect the average \textit{PerfScore} over the course of these iterations.  As a result, we rank the environmental variables based on the disparity between the \textit{PerfScore} at the minimum value and the \textit{PerfScore} at the maximum value. The environments associated with the variables exhibiting the greatest discrepancy in \textit{PerfScore} can be considered the most challenging scenarios for the DRL model.

We make the assumption that there exists a linear relationship between the change in environment variables and \textit{PerfScore}. Consequently, we can approximate the decrease or increase in \textit{PerfScore} with respect to changes in the environment variable. This assumption will be valuable for designing an evolutionary environment that can effectively regulate the model's \textit{PerfScore} at a specific level. In the experiment section, we will provide evidence to support the claim that the relationship between \textit{PerfScore} and environmental variables is indeed close to a linear relationship.

\vspace{-0.3cm}

\subsection{\textbf{Evolutionary Curriculum Training Environment}} \label{sec:evolutionary_training}

We employ a curriculum training approach that strategically focuses on tackling the most arduous environments for the DRL model. Our objective is to enhance the model's performance in these challenging scenarios by deliberately exposing it to the most demanding environments, intentionally inducing failures, and reducing \textit{PerfScore}. By doing so, we aim to ensure that the model's failures occur specifically within the environment we seek to address and overcome. To accomplish this, we systematically reduce the difficulty level of all other variables in the training environment, effectively minimizing the likelihood of failures resulting from factors unrelated to the specific challenges we aim to tackle.

Once the model's \textit{PerfScore} reaches a threshold of 0.75 in the current environment, the training environment adjusts the most challenging variable to increase the corresponding environment's difficulty. This exposes the DRL model to a more challenging scenario while simultaneously regulating the model's \textit{PerfScore} within a certain range. This iterative process continues until the most challenging variable reaches its maximum level within its range. Once this maximum level is reached, signifying the completion of a curriculum iteration, the training environment adjusts all other variables to increase the overall difficulty of the environment. It is worth mentioning that we initially attempted to implement a pipeline similar to the evaluate-train-evaluate-train approach to ensure that the model will be trained against the most challenging environment. However, this method significantly slowed down the training process. Consequently, informed by empirical observations, we chose to train within the most demanding environment for two curriculum iterations before transitioning to the next most challenging environment.

%% file: TextFiles/Experiments.tex
\section{\textbf{Training and Experiments}} \label{sec:experiments}

\subsection{\textbf{Finetuning GA3C-CADRL model}} \label{R-score}

In this section, we will fine-tune an open-source DRL model named GA3C-CADRL [GA3C-CADRL-10-LSTM-4] \cite{Everett_Motion} using our Evolutionary Curriculum Training approach.

\begin{table}[h]
\centering
\resizebox{0.5\textwidth}{!}
{
\begin{tabular}{@{}l|c@{}}
\toprule
Variables & Difference Between Min and Max\\ \toprule
Room Number & 0.6093 \\
Ped's Policy & 0.2759 \\
Ped's Number & 0.2166 \\
Ped's Speed & 0.1753 \\
Room Size & 0.0509 \\
Corridor Width & 0.0074 \\
Convexity Level & 0.0064 \\ \bottomrule
\end{tabular}
}
\caption{The relational expression between \textbf{PerfScore} and different variables. Those functions are used to evaluate and generate the next step of the training process.}
\label{tab:R-score}
\end{table}

\begin{figure*}[ht]
    \centering
    \includegraphics[width=\textwidth]{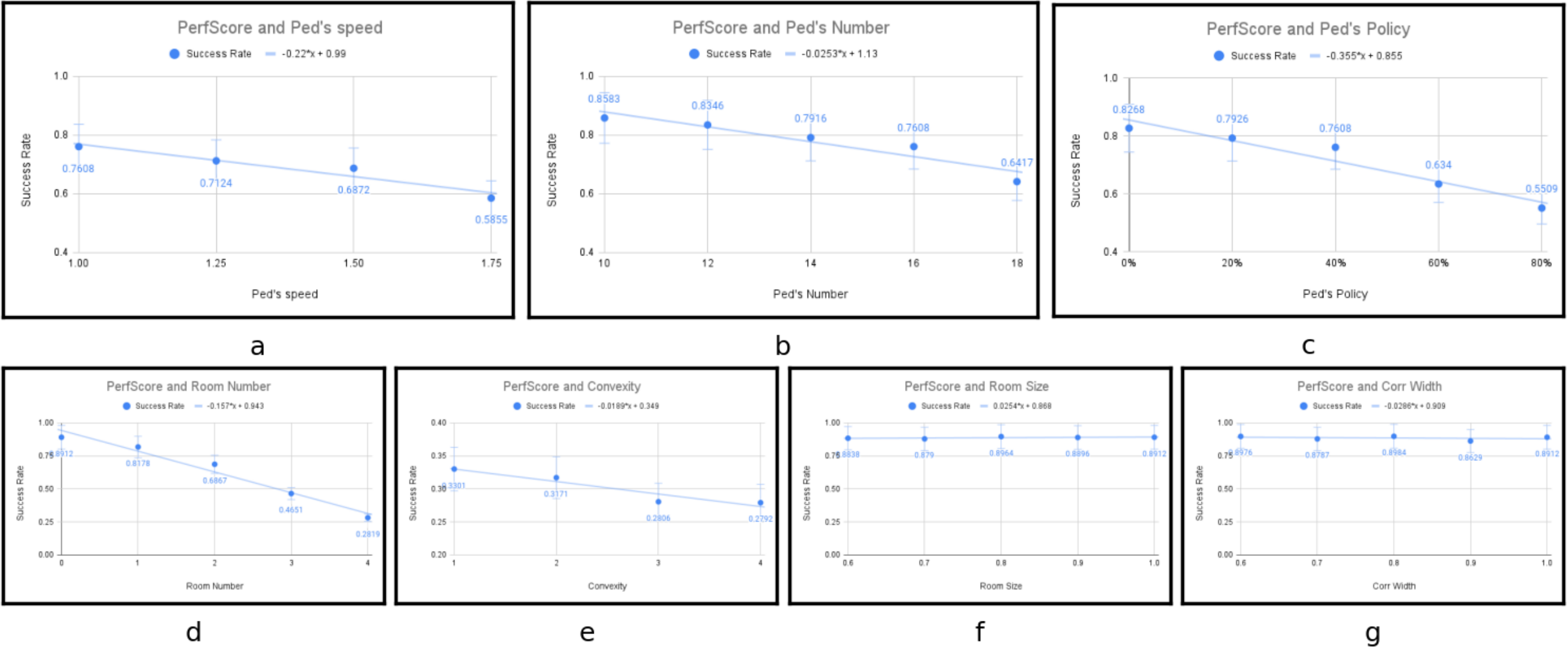}
    \caption{These graphs correspond to each variable's PerfScore. The points in the graph represent the PerfScore corresponding to a certain variable level. The equation of each fitted line showed under the graph title.}
    \label{fig:plot}
\end{figure*}
\vspace{-0.3cm}

Firstly, we evaluate the pretrained GA3C model using our evaluation process. The ranking table \ref{tab:R-score} clearly illustrates that the GA3C model performs poorly in complicated structured maps and also exhibits sub-optimal performance when encountering pedestrians with a hard policy, such as random walking or circle walking. These findings align precisely with the training environment of GA3C, which primarily consists of wide open spaces with pedestrians following easier policies such as linear walking or remaining stationary. To evaluate this model, we use the \textit{PerfScore} in section \ref{sec:perf_score}. In this section \ref{sec:doability}, we prove that the \textit{PerfScore} can be used to evaluate the success rate of each variable.

This clearly demonstrates how the evaluation method can help DRL models identify their limitations and facilitate the provision of appropriate curriculum in the Evolutionary Training Environment.

During the Evolutionary Training Process, the training environment effectively regulates the \textit{PerfScore} within a reasonable range. Additionally, it is noteworthy that the DRL model successfully completes training for the curriculum in a relatively short period of time (45000 iterations). This observation may suggest that our conservative approach in selecting the reasonable range for variables could be further explored. In the forthcoming experiment section, we will employ our newly trained \textit{Evolutionary-GA3C} model to conduct experiments in diverse environments with more radical variable setups. This will provide an opportunity to demonstrate the enhanced generality achieved through our Evolutionary Curriculum Training.

To further validate our assumption regarding the linear relationship between the environment variable and \textit{PerfScore}, we carried out a more comprehensive evaluation process by employing a greater number of sampling points. For each environment variable, we initially fixed the other variable at a moderate level and uniformly sampled five points within its reasonable range. Subsequently, we generated 500 maps corresponding to all the sampled points and performed 5000 simulations for each setup, just like in our method. We then collected the final average \textit{PerfScore} as observed in our evaluation process. To further analyze the data, we fitted polynomial curves on the data points using these five points per variable like fig. \ref{fig:plot}. Notably, it was unsurprising to find that the relationship between all seven variables and the \textit{PerfScore} was closely approximated by linear patterns. Moreover, the two variables identified in the table \ref{tab:R-score} as having minimal impact on the \textit{PerfScore} exhibited nearly flat lines in the fitted curves.

Since all of our evaluations involve only one variable and \textit{PerfScore}, one might question the impact of simultaneously changing multiple variables during each curriculum training. Concerns may arise regarding the potential for excessive \textit{PerfScore} decreases when more than one variable is modified concurrently.

First and foremost, based on our observations, we have indeed noticed a doubling effect in certain pairs of variables, such as Pedestrian policy and Pedestrian speed. However, considering that we have collected all the variable-\textit{PerfScore} relations under moderate difficulty and have adopted a conservative approach in defining the reasonable range, the current problem does not render the training environment incapable of regulating the \textit{PerfScore} effectively. Nevertheless, when employing a more radical variable range setup, this effect must be taken into account and carefully considered.

\subsection{\textbf{Hierachical Navigation Framework}} \label{sec:framework}

To experimentally validate our waypoint planner, we implemented our navigation framework through the ROS-based \cite{quigley_ros_2009} STDR \cite{tsardoulias_stdr_simulator_2014}, or Simple Two-Dimensional Robots, simulation environment. Preliminary experiments were performed in ROS Noetic on an Ubuntu 20.04 machine using the {\fontfamily{qcr}\selectfont move\_base} ROS package \cite{ros_move_base}. To evaluate the effectiveness of our proposed strategy, which combines the waypoint planner and our baseline RL model, in static structured environments, dynamic environments, and complex structured environments with dynamic obstacles, we conducted three different experiments in the ROS-based \cite{quigley_ros_2009} STDR \cite{tsardoulias_stdr_simulator_2014}. To integrate the CADRL ROS node \cite{Everett_Motion} with our \textit{Evolutionary GA3C-CADRL} model, we made modifications that enabled the nodes to work together in the ROS environment. Specifically, we adjusted the input of the CADRL ROS node to match the output from our waypoint planner. For more detail, check the appendix \ref{sec:local_waypoint_planner}.

\subsection{\textbf{Experiments Setup}}

In this section, we will test our \textit{Evolutionary-GA3C} model and hierarchical model in three different environments.

\begin{itemize}

\item  \textbf{Static Environment:} We want to evaluate that our Evolutionary Training Environment has great improvements for agents in static structured environments. We used 10 maps in Figure \ref{fig:map} with various room numbers and convexity levels. For each map, we want our agents to traverse the longest path within each map. During each navigation, the robot will be influenced by the corridors and corners of the rooms during cornering or U-turning, which causes collisions and stagnation.

\begin{figure}[ht]
    \centering
    \includegraphics[width=0.5\textwidth]{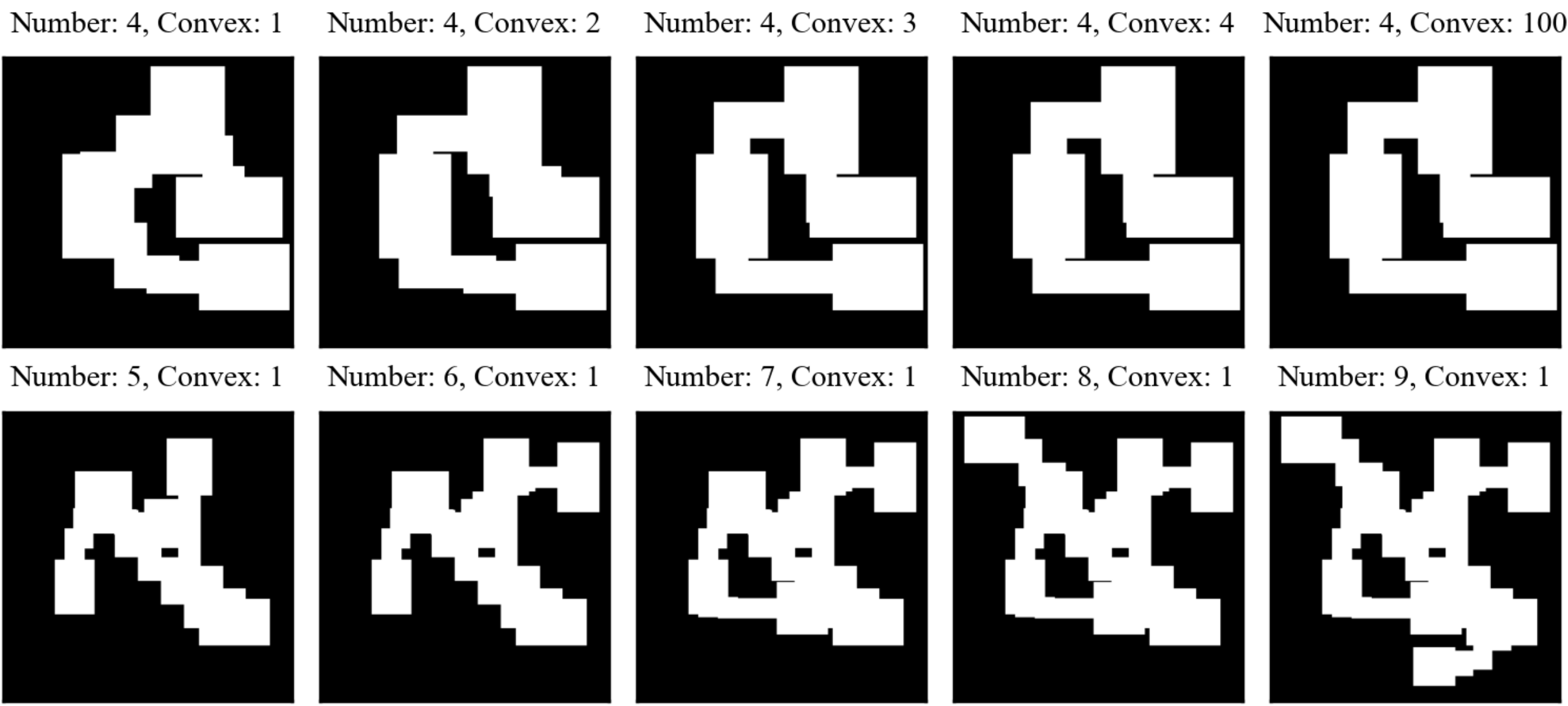}
    \caption{We have ten randomly generated maps with room numbers from 4 to 9 and Convex levels 1, 2, 3, 4, and 100. We let the robot navigate through the longest path on every map.}
    \label{fig:map}
\end{figure}

\item  \textbf{Dynamic Environment:} We wanted to prove that our random policy is useful for avoiding colliding with pedestrians. We set up 20, 30, or 40 randomly generated dynamic obstacles in an empty room. Those dynamic obstacles have various random velocities, waypoints, and radii, which can be used to simulate real pedestrians. The example is shown in Figure \ref{fig:random}.

\begin{figure}[ht]
    \centering
    \includegraphics[width=0.30\textwidth]{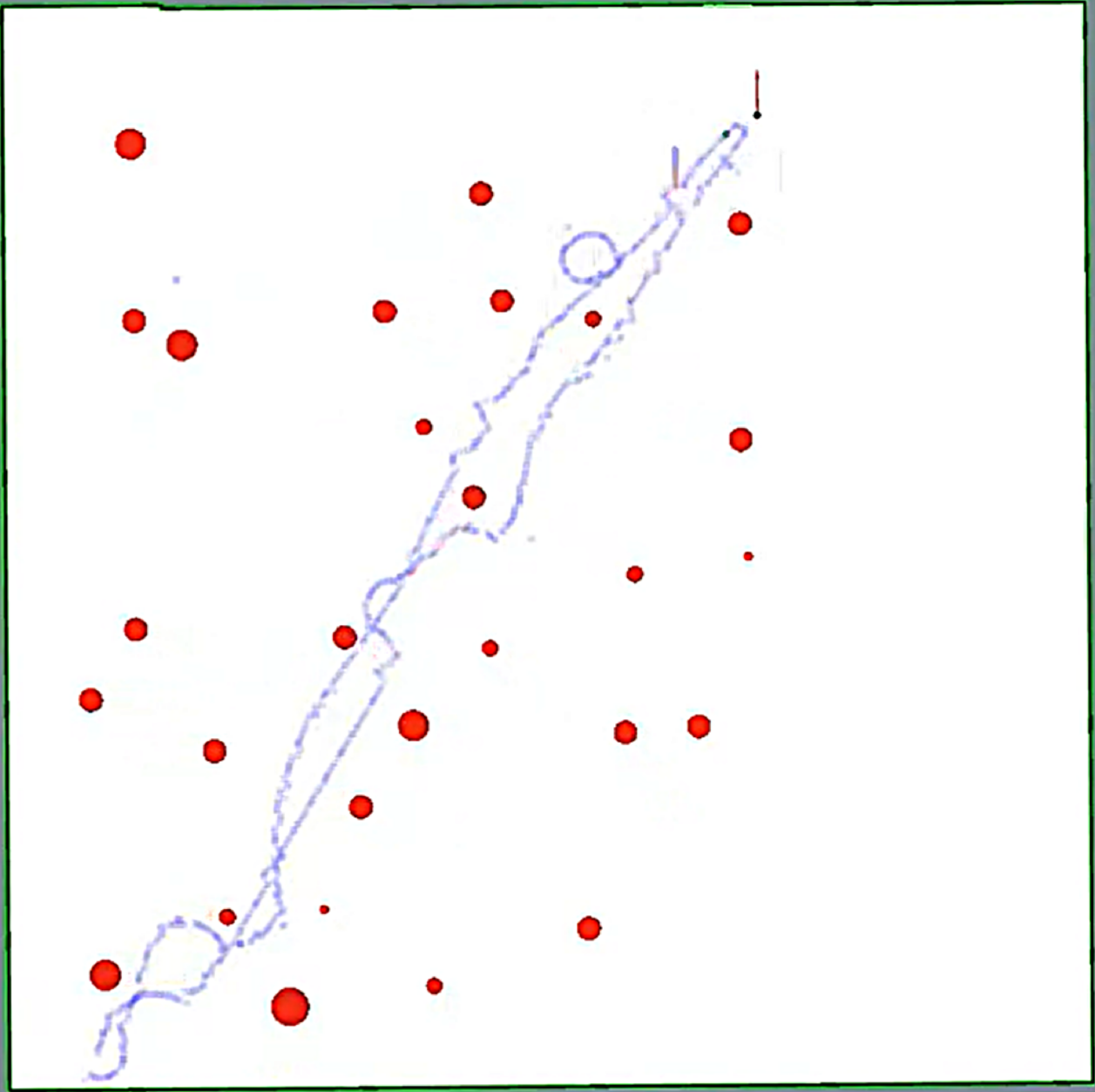}
    \caption{Example environment with 30 randomly generated agents (red) with different velocities and radii. The blue lines show the trajectory of the robot (green), which navigates across the dynamic obstacles.}
    \label{fig:random}
\end{figure}

\item  \textbf{Complex Environment:} We built five different Maps shown in Figure \ref{fig:Maps}. We have four types of maps: one has randomly generated squares and points named Campus, one has only random points named Forest, two generated from Evolutionary Room/Maze Generator in Section \ref{sec:environment_setup} with room numbers 9 and 8 named Room1 and Room2, and one Irregular Room contains obstacles with irregular shapes. In each map, we will generate 10-20 agents with various random velocities and radii in different locations. Those agents will move in a cycle between two points. Moreover, these agents may suddenly drill out of the wall, which the sensor cannot detect, to test the speed and generalization ability of our model. 15 success examples are shown in Figure \ref{fig:Successful Cases}.

\begin{figure}[h]
    \centering
    \includegraphics[width=0.5\textwidth]{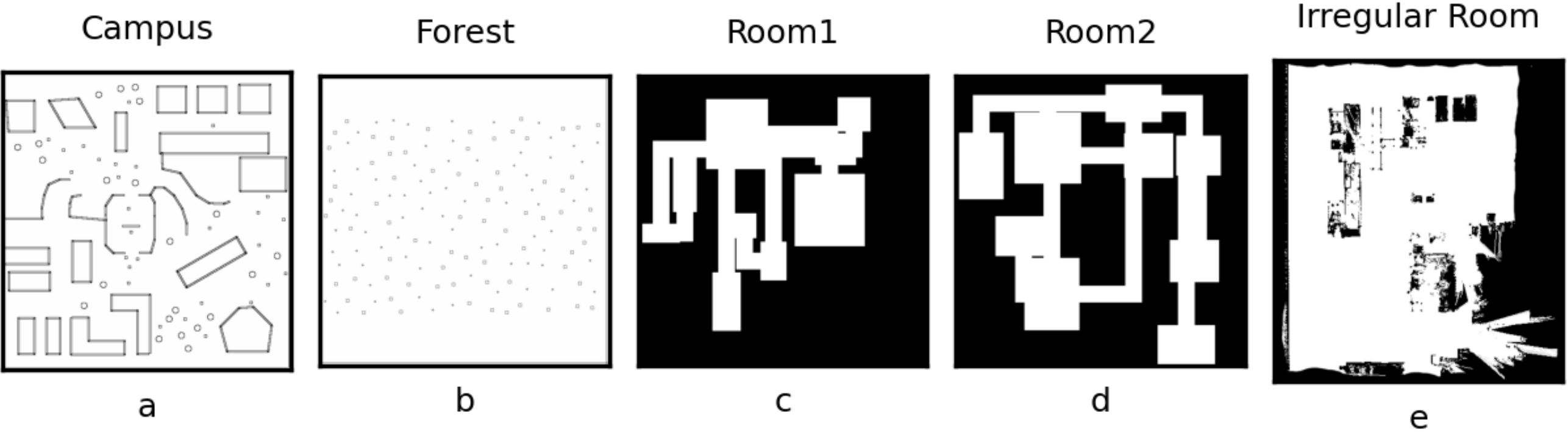}
    \caption{Five different maps used in experiments.}
    \label{fig:Maps}
\end{figure}

In each map, we set up three goals for our agents to walk. Each goal contains different static obstacles and moved agents. To evaluate our models, we will let each of them go through each goal 10 times and then calculate their success rates and number of collisions.



\end{itemize}

\subsection{\textbf{Experiments Analysis}}

\paragraph{\textbf{Static Environment}}
Within this experimental investigation, we systematically assess the efficacy of our novel Evolutionary Training Environment within static structured environments. To gauge its performance, we establish a comparative framework involving several baseline models, namely waypoint planner + CADRL\cite{DBLP:journals/corr/ChenLEH16}, waypoint planner + GA3C-CADRL\cite{Everett_Motion}, and our own waypoint planner + \textit{Evolutionary GA3C-CADRL}. Subsequently, we test each model by navigating through the longest paths present within ten distinct maps, repeating the process ten times to ensure statistical robustness.

Since the RL models are not sensitive enough for static scenes, they usually hit the wall and stall, causing navigation failure. Hence, we will evaluate the success rate for each of the models and evaluate their performance. The results are shown in the table \ref{tab:static-table}.

According to the result, we can see that our robot trained with an Evolutionary Training Environment outperforms the other models. The success rate of the proposed approach is 82\%, whereas the second highest rate is 60\%. Previous models were prone to crashing into walls and terminating motion when cornering or U-turning, while our model can reduce these occurrences. (more cases can be found in examples on our \href{https://sites.google.com/view/hierarchical-navigation}{GoogleSite})

\begin{table}[h]
\centering
\resizebox{0.5\textwidth}{!}
{
\begin{tabular}{@{}llccc@{}}
\toprule
Room & Convex & WP+CADRL & WP+GA3C & \textbf{WP+Evolutionary-GA3C}\\ \toprule
4 & \textbf{1} & 7/10 & 7/10 & \textbf{9/10}\\
4 & \textbf{2} & 7/10 & 6/10 & \textbf{10/10} \\
4 & \textbf{3} & 6/10 & 6/10 & \textbf{7/10} \\
4 & \textbf{4} & 7/10 & 6/10 & \textbf{8/10} \\
4 & \textbf{100} & 6/10 & 6/10 & \textbf{8/10} \\ \toprule
\textbf{5} & 1 & 5/10 & 6/10 & \textbf{8/10} \\
\textbf{6} & 1 & 6/10 & 6/10 & \textbf{9/10} \\
\textbf{7} & 1 & 5/10 & 5/10 & \textbf{7/10} \\
\textbf{8} & 1 & 5/10 & 7/10 & \textbf{9/10} \\
\textbf{9} & 1 & 4/10 & 5/10 & \textbf{7/10} \\ \toprule
Overall & & 55\% & 60\% & \textbf{82\%} \\ \bottomrule
\end{tabular}
}
\caption{Results of baseline methods against waypoint planner (WP) + \textit{Evolutionary-GA3C} in static environments. It includes 10 different maps with different room numbers and complexity levels. Here are the success rates of each map for 10 navigations.}
\label{tab:static-table}
\end{table}

\paragraph{\textbf{Dynamic Environment}}
In this experiment, we evaluate the effectiveness of our proposed Evolutionary Training Environment in an empty room with a lot of randomly generated dynamic obstacles. We compare our model with some baseline models: waypoint planner + CADRL\cite{DBLP:journals/corr/ChenLEH16}, waypoint planner + GA3C-CADRL\cite{Everett_Motion}, and our waypoint planner + \textit{Evolutionary GA3C-CADRL}. We randomly generated 20 dynamic obstacles, 30 dynamic obstacles, and 40 dynamic obstacles with various velocities, waypoints, and radii in an empty map for each navigation.

To evaluate the performance, we will collect the total collision number during 20 navigation with different randomly generated obstacles. The results are shown in the table \ref{tab:dynamic-table}.

According to the result, we can see that our robot trained in an Evolutionary Training Environment is better than the other baseline models. The average total collision number is 11.67, whereas the other two models are 18.67 and 27. It proves that our random policy is helpful in complex dynamic environments.

\begin{table}[h]
\centering
\resizebox{0.5\textwidth}{!}
{
\begin{tabular}{@{}lccc@{}}
\toprule
Num of Agents & WP+CADRL & WP+GA3C & \textbf{WP+Evolutionary-GA3C}\\ \toprule
20 Agents & 20 & 15 & \textbf{7}\\
30 Agents & 25 & 19 & \textbf{12} \\
40 Agents & 36 & 22 & \textbf{16} \\ \toprule
Overall & 27 & 18.67 & \textbf{11.67} \\ \bottomrule
\end{tabular}
}
\caption{Results of baseline methods against waypoint planner (WP) + \textit{Evolutionary-GA3C} in dynamic environments. It includes 20, 30, or 40 random dynamic obstacles in an empty room. We calculate the total collision number during 20 navigations.}
\label{tab:dynamic-table}
\end{table}

\vspace{-0.4cm}

\paragraph{\textbf{Complex Environment}}
In this experiment, we evaluate the effectiveness of our proposed model waypoint planner + \textit{Evolutionary GA3C-CADRL}. We compare our framework with five baseline models of CADRL \cite{DBLP:journals/corr/ChenLEH16}, \textit{Evolutionary GA3C-CADRL}, waypoint planner + MPC, waypoint planner + CADRL, and waypoint planner + GA3C-CADRL \cite{Everett_Motion}. The CADRL model is a trained RL model without a global planner or any map information. We aim to prove that the waypoint planner is helpful in a complex structured environment. The waypoint planner + MPC model is a traditional model-based algorithm with waypoint planner and map information. We show that the RL-based model can outperform the traditional model-based multi-agent algorithm in a complex environment. The other three models are all RL-based models with waypoint planner and map information. We observe that our Evolutionary RL framework is better than the other RL policies.

To assess the performance of each model, we propose to let the agent with different trained models walk ten times for each goal shown in section V.C.1. Then, we set up the limit time to 30 seconds. If the agent can reach the goal in 30s, it is a successful navigation. Otherwise, it fails. Then, we will calculate the success rate for each model. Also, we measure the number of collisions that occur between agents as well as between agents and static obstacles. For each goal, we calculate the total number of collisions in 10 times of navigation. If our agent is too close to other agents or obstacles, we will count it as one collision.

\begin{figure*}[htb!]
    \centering
    \includegraphics[width=0.99\textwidth]{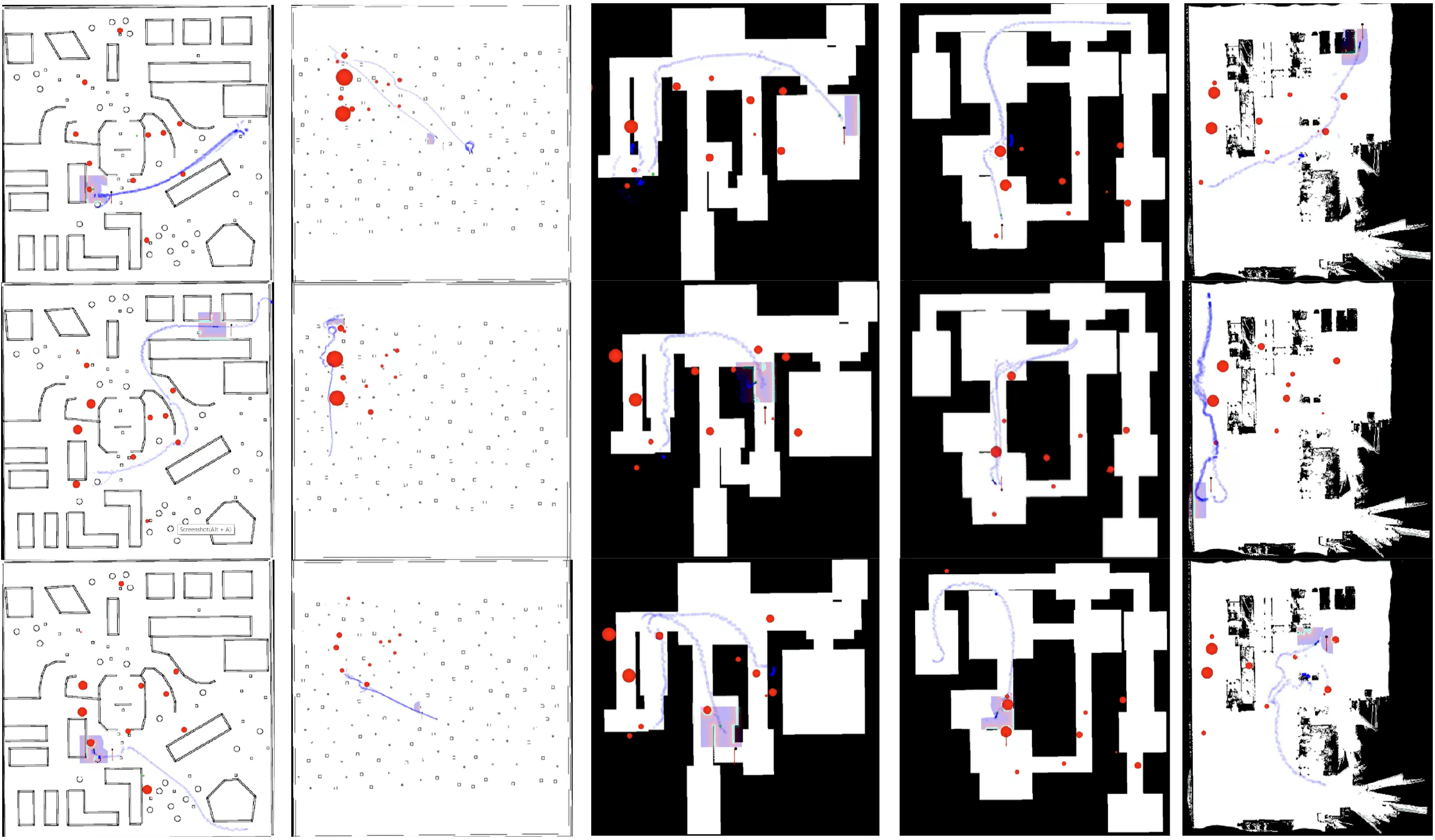}
    \caption{15 successful examples with our waypoint planner + \textit{Evolutionary-GA3C} model in 15 goals from five different maps. \\
    In the map, the blue line shows the successful path, the blue area is the detection area of the laser scan, the point inside the blue area is the robot, the other red points are moved agents, and the point with a "tail" in front of the robot is the waypoint.}
    \label{fig:Successful Cases}
\end{figure*}

Then, we start to evaluate those five different models with 15 different goals from five different maps. The results of all the models are shown in Table \ref{tab:my-table}.  The successful cases with our waypoint planner + \textit{Evolutionary GA3C-CADRL} model for all those 15 goals are visualized in Figure. \ref{fig:Successful Cases}. Everything, such as the agent, the dynamic obstacles, the waypoints, and the map information are visualized in them. More examples and comparisons can be found in our \href{https://sites.google.com/view/hierarchical-navigation}{GoogleSite}.

\begin{table*}[h]
\centering
\resizebox{0.8\textwidth}{!}
{
\begin{tabular}{@{}llcc|cccc@{}}
\toprule
Map & Path & CADRL & \textbf{Evolutionary-GA3C} & WP+MPC & WP+CADRL & WP+GA3C & \textbf{WP+Evolutionary-GA3C}\\ \toprule
Campus & 1 & 30\% / 15 & 60\% / 9 & 70\% / 37 & 90\% / 12 & 90\% / 7 & \textbf{100\% / 3}\\
       & 2 & 0\% / 21 & 0\% / 13 & 30\% / 45 & 70\% / 19 & \textbf{80\% / 17} & 80\% / 18 \\
       & 3 & 0\% / 18 & 40\% / 10 & 80\% / 39 & 60\% / 15 & 70\% / 13 & \textbf{80\% / 6} \\  \toprule
Forest & 1 & 60\% / 10 & 80\% / 6 & 100\% / 31 & 90\% / 10 & 100\% / 9 & \textbf{100\% / 4}\\
       & 2 & 80\% / 4 & 100\% / 1 & 100\% / 10 & 100\% / 3 & 100\% / 3 & \textbf{100\% / 0} \\
       & 3 & 80\% / 8 & 90\% / 3 & 100\% / 13 & 100\% / 7 & 100\% / 5 & \textbf{100\% / 0} \\ \toprule
Room1 & 1 & 0\% / 13 & 20\% / 10 & 0\% / 20 & 50\% / 17 & 50\% / 13 & \textbf{80\% / 9} \\
      & 2 & 0\% / 10 & 30\% / 10 & 0\% / 20 & 40\% / 11 & 60\% / 9 & \textbf{70\% / 11} \\
      & 3 & 0\% / 10 & 30\% / 11 & 0\% / 20 & 50\% / 15 & 50\% / 14 & \textbf{80\% / 10}  \\ \toprule
Room2 & 1 & 0\% / 20 & 30\% / 10 &  20\% / 35 & 70\% / 10 & 70\% / 10 & \textbf{90\% / 9}  \\
      & 2 & 0\% / 20 & 20\% / 11 & 10\% / 25 & 60\% / 7 & 70\% / 8 & \textbf{80\%} / 9  \\
      & 3 & 0\% / 20 & 30\% / 8 & 0\% / 33 & 50\% / 16 & 60\% / 14 & \textbf{70\% / 9}  \\ \toprule
Irregular & 1 & 0\% / 10 & 20\% / 12 & 70\% / 43 & 90\% / 13 & 90\% / 10 & \textbf{100\% / 4} \\
             room & 2 & 0\% / 10 & 30\% / 8 & 70\% / 51 & 50\% / 19 & 70\% / 17 & \textbf{90\% / 7} \\
               & 3 & 40\% / 16 & 80\% / 3 & 100\% / 29 & 70\% / 13 & 90\% / 10 & \textbf{100\% / 0} \\ \toprule
Overall &  & 19.3\% / 13.7 & 44\% / 8.3 & 50\% / 30.1 & 69.3\% / 12.5 & 76.7\% / 10.6 & \textbf{87.3\% / 6.6} \\ \bottomrule
\end{tabular}
}
\label{tab:my-table}
\vspace{0.2cm}
\caption{Results of baseline methods against waypoint planner (WP) + \textit{Evolutionary-GA3C}. It includes five different maps and 15 different goals. The first column of each method shows the success rates in ten navigation, and the second column shows the total amount of collisions in ten navigation. The left side of the vertical line is the version without Waypoint Planner, and the right side has the Waypoint Planner}
\end{table*}

Now, we provide key insights from our obtained results:
\begin{itemize}

\item For the RL-based model without waypoint planner and map information (CADRL), we find that it cannot finish most tasks except the task in the forest. The average success rate and collision counts are only 19.3\% and 13.7 collisions. For our \textit{Evolutionary GA3C} model, although it's much better than the original CADRL model because it can handle more complex information in a structured environment and dynamic obstacles, the success rate is still only 44\%. Without the waypoint planner and map information, CADRL cannot find a doable path at once, and it should be pretty slow for it to work in a complex structured environment. Also, sometimes it gets stuck in some impassable dead end and stops moving. The reason it works fine in forest maps is that those small cycles can be treated as static pedestrians, so the map information doesn't matter. Through this comparison, we find that the waypoint planner is very important in complex structured environments.

\item For the model-based MPC model with waypoint planner, we find that it is worse than other RL-based models with waypoint planner. Through the experiments we found the success rate to be lower than the other RL-based model, and the collision number to be significantly larger. During the experiment, we find that the MPC model in our direction often directly collides with the other moving agents because it is not fast enough. Also, it is often impossible to plan the route again after being hit. Through this comparison, we find that the RL-based model is better than the model-based MPC model as a local planner in complex structured environments.

\item For RL-based models incorporating a waypoint planner, they consistently demonstrate outstanding performance in complex structured environments with numerous dynamic agents. Notably, our proposed \textit{Evolutionary GA3C-CADRL} model outperforms other baseline models in terms of both success rate and collision frequency in our experiments. In the absence of the waypoint Planner, our model achieves a success rate 24.7\% higher than the CADRL model. When the waypoint Planner is employed, our model achieves a success rate 9.7\% higher than the WP+GA3C model while reducing four average collisions for every ten navigations. Additionally, our method exhibits significant improvements in environments with many corners like room1 and room2, demonstrating its capability to handle complex environments beyond the capabilities of prior methods. These results collectively support the effectiveness of our Evolutionary Training Environment in addressing the challenges posed by complex structured environments with dynamic obstacles.

\end{itemize}

%% file: TextFiles/Conclusion.tex
\section{\textbf{Conclusion}} \label{sec:conclusion}
This proposed work presents a hierarchical navigation framework with a gap waypoint planner and Evolution-GA3C Reinforcement Learning Algorithm to help the robot work in a complex structured environment with static obstacles and moving pedestrians and agents. It also proposes a novel training insight, which is able to conduct collision avoidance training in complex environments without map information. The new approach outperforms other baseline methods in different maps and goals with great generalization ability.

In the future, we will try to add more maps or agent policies and use neural networks to evaluate the difficulty level of the training environment during the training process more intelligently. Also, we expect that the hierarchical framework trained from our Evolutionary Environment can be used on real robots.

%% file: TextFiles/Appendix.tex
\section{\textbf{Appendix}}
\subsection{\textbf{Problem Fomulation}} \label{sec:problem_formulation}
First, we summarize the multi-agent collision avoidance task we are exploring with this proposed work. This task concerns the collision-free navigation of an ego-robot $\mathcal{R}_{ego}$ through partially unobservable environments. These environments are comprised of a set of static obstacles $\mathcal{O}_{static}$ such as corridors, tables, and walls and a set of dynamic agents $\mathcal{O}_{dynamic}$. Together, these comprise the complete set of obstacles $\mathcal{O} = \mathcal{O}_{static} \cup \mathcal{O}_{dynamic}$. The ego-robot is to travel from its starting pose $\mathbf{x_{0}} = [p_{x,0}, p_{y, 0}, \psi_{0}]$ to its global goal pose $\mathbf{x_{GG}} = [p_{x, GG}, p_{y, GG}, \psi_{GG}]$ without any collisions. In order to navigate, we assume the ego-agent has access to a noiseless $360\degree$ laser scan $\mathcal{L}$ with some maximum range $r_{max}$. This laser scan returns finite range values at the bearings where obstacles from $\mathcal{O}$ are within the sensing range.


For the learned collision avoidance model GA3C-CADRL, we refer readers to \cite{Everett_Collision} for complete details, but we will provide a brief description of relevant inputs and outputs here. The overall state at timestep $t$ is split into an observable portion $\mathbf{s^o_t}= [p_{x}, p_{y}, v_{x}, v_{y}, r] \in \mathbb{R}^5$ comprised of an agent's position, velocity, and radius and a hidden portion $\mathbf{s^h_t} = [p_{x, LG}, p_{y, LG}, v_{pref}, \psi]^T \in \mathbb{R}^4$ comprised of the local goal (or local waypoint) position as well as the preferred speed and current orientation of the robot, making $\mathbf{s_t} = [\mathbf{s^o_t}, \mathbf{s^h_t}]^T \in \mathbb{R}^9$. The ego-agent's state vector is represented as $\mathbf{s_t}$, and the state vector $\mathbf{\Tilde{s_t}}$ represents the state of other proximal agents (the number of which varies). The collision avoidance policy $\pi : (\mathbf{s_t}, \mathbf{\Tilde{s_t}^o}) \rightarrow \mathbf{u_t}$ acts on the entire ego-robot state and outputs a speed and a heading angle, $\mathbf{u_t} = [v_t, \psi_t] \in \mathbb{R}^2$. Given that the velocities of other agents are not directly observable, a Kalman filter is deployed to estimate such values.


\subsection{\textbf{Hierachical Navigation Framework Diagram}}
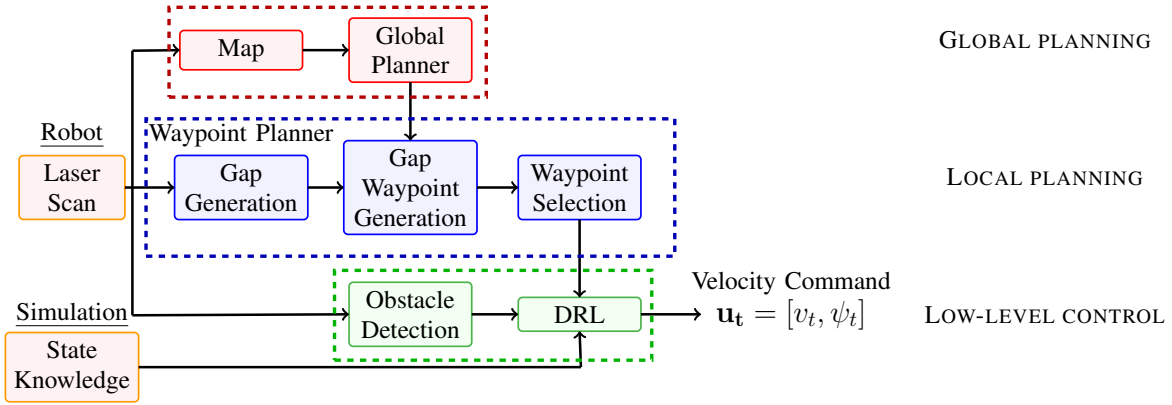
\begin{figure*}[t]
  \vspace*{0.06in}
  \centering
  \hspace*{-0.62in}
  \begin{tikzpicture}[inner sep=0pt,outer sep=0pt]
    \node[anchor=south west] at ($(0, 0)$)
    	{{\scalebox{0.8}{\input{TextFiles/diagram.tex}}}};
  \end{tikzpicture}%
  \caption{Hierachical navigation diagram including global planner, local waypoint planner, and local collision avoidance model. The collision avoidance model outputs the desired speed $v_t$ and heading $\psi_t$.}
  \label{fig:diagram}
  \vspace*{-1.1em}
\end{figure*}

The fig. \ref{fig:diagram} shows our entire Hierarchical Navigation Framework, which contains a high level Global Planner, intermediate-level Gap Waypoint Planner, and low-level DRL-based collision avoidance model.

\subsection{\textbf{Local Waypoint Planner}} \label{sec:local_waypoint_planner}
To obtain local navigation waypoints, we incorporate a gap-based local planner into our hierarchical navigation framework. Gap-based planners operate on perception data to generate robot-centric gaps, or regions of free space, that are formed between the leading and trailing edges of obstacles. More information regarding how these gaps are formed can be found from the cited work \cite{Xu_Potential}, but here we will briefly detail the gap detection algorithm used to obtain local waypoints.

\subsubsection{Gap Generation}

During iteration through the scan $\mathcal{L}$, gaps are detected from two situations: 
\begin{enumerate}[label=(\alph*)]
    \item Angular interval of free space that robot can fit through: 
    \[I = [i, i + k] \hspace{0.25cm} \text{s.t.} \hspace{0.25cm} \forall j \in I, \hspace{0.25cm}\]  
 \[\mathcal{L}(j) = r_{max}, \hspace{0.25cm} \text{and} \hspace{0.25cm} \| \zeta_i - \zeta_{i+k} \| > 2r_{robot} \]
    \item Instantaneous change in range that suggests free space that robot can fit through: \[\lvert \mathcal{L}(i+1) - \mathcal{L}(i) \rvert > 2r_{robot} \]
\end{enumerate}
where $r_{robot}$ is the radius of the robot. Gaps that satisfy the first situation are classified as \say{swept} gaps with high line-of-sight visibility, and gaps that satisfy the second situation are classified as \say{radial} gaps with poor visibility. This pass through the scan $\mathcal{L}$ provides us with a set of \say{raw} gaps $\mathcal{G}_{raw}$.

Within the set of gaps $\mathcal{G}_{raw}$, those that are classified as radial gaps can potentially be merged together to form additional swept gaps. 
This pass through $\mathcal{G}_{raw}$ generates a set of simplified gaps $\mathcal{G}_{simp}$. From $\mathcal{G}_{simp}$, a set of steps are performed to to ensure high line-of-sight visibility during gap passage and account for the robot radius by inflating the gap sides, leaving us with a final set of manipulated gaps $\mathcal{G}_{manip}$.

\subsubsection{Gap Waypoint Generation and Selection}
Once the manipulated gaps are generated, a waypoint is placed within each candidate gap. Waypoints are placed within each gap to allow us to evaluate passing through each one of the possible manipulated gaps. This waypoint planner acts as the lower level of a hierarchical global-local navigation planner, so the waypoint generation step extracts the local segment of the current global plan and extracts the furthest pose in that segment as a \say{global plan goal}. 


If the global plan goal lies within one of the gaps, that gap's waypoint is set equal to the global plan goal. Otherwise, the waypoints of the gaps are biased towards whichever side (left or right) the global plan goal falls on. From the set of manipulated gaps, we obtain a set of local gap waypoints, $G$. One of the waypoints from $G$ is picked based on a user-defined cost function which factors in distance to global goal as well as distance to obstacles, and the lowest cost waypoint $g$ is selected to pass to the obstacle avoidance model.


\subsection{\textbf{Evaluate doability of PerfScore}} \label{sec:doability}
In this section, since the \textbf{PerfScore} in the training process may not be exactly the same as the real success rate, we want to prove that using the \textbf{PerfScore} to classify the difficulty level is doable. We want to do an experiment in the ROS environment to prove it. The ROS environment setup is the same as the section \ref{sec:framework}.

\subsubsection{Experiments} 
In sec. \ref{sec:environment_setup}, the concept of difficulty level is delineated into static and dynamic components. Within the static difficulty level, the influential factors encompass the room number and convexity level, while the size of rooms and corridors remains inconsequential. On the other hand, the dynamic difficulty level pertains to the impact of dynamic obstacles, including their speed, quantity, and governing policy. Considering our objective of demonstrating the accurate reflection of difficulty through the \textbf{PerfScore}, we have elected to focus on the static difficulty aspect. This choice enables a simpler design for evaluating our methodology.

We generated 16 maps, employing a systematic approach of varying factors within the static difficulty level. Then, we set up experiments in STDR Environments shown in section \ref{sec:framework} Specifically, for each of the factors, namely room number, convexity level, room size, and corridor width, we generated four distinct maps while maintaining consistency across the remaining factors. These maps follow a similar structure as depicted in Figure \ref{fig:comparasion}, but with modifications in room and corridor dimensions. Subsequently, we conducted 50 navigation trials wherein the robot traversed the longest path within each map, allowing us to ascertain the success rates. The findings are presented in Table \ref{tab:Difficulty-table}. The observed success rate trends align closely with the outcomes derived from the \textbf{PerfScore} analysis, affirming that room number exerts a more pronounced influence compared to convexity level, while room size and corridor width demonstrate almost zero impact.

\begin{table}[h]
\centering
\resizebox{0.5\textwidth}{!}
{
\begin{tabular}{@{}llllc@{}}
\toprule
Num of Rooms & Room Size & Corridor Width & Convexity & Success Rate\\ \toprule
4 & \textbf{0.7} & 0.5 & 1 & 0.94 \\
4 & \textbf{0.8} & 0.5 & 1 & 0.90 \\
4 & \textbf{0.9} & 0.5 & 1 & 0.94 \\
4 & \textbf{1.0} & 0.5 & 1 & 0.92 \\ \toprule
4 & 0.8 & \textbf{0.5} & 1 & 0.94 \\
4 & 0.8 & \textbf{0.6} & 1 & 0.90 \\
4 & 0.8 & \textbf{0.7} & 1 & 0.92 \\
4 & 0.8 & \textbf{0.8} & 1 & 0.90 \\ \toprule
4 & 0.8 & 0.5 & \textbf{1} & 0.94 \\
4 & 0.8 & 0.5 & \textbf{2} & 0.92 \\
4 & 0.8 & 0.5 & \textbf{3} & 0.88 \\
4 & 0.8 & 0.5 & \textbf{4} & 0.84 \\ \toprule
\textbf{4} & 0.8 & 0.5 & 1 & 0.94 \\
\textbf{5} & 0.8 & 0.5 & 1 & 0.90 \\
\textbf{6} & 0.8 & 0.5 & 1 & 0.84 \\
\textbf{7} & 0.8 & 0.5 & 1 & 0.78 \\ \bottomrule
\end{tabular}
}
\caption{Change of success rate when we change the factors of static difficulty levels. Collect the success rate during 50 navigations for each map.}
\label{tab:Difficulty-table}
\end{table}

%% file: TextFiles/diagram.tex
\tikzstyle{block} = [draw, rectangle, text centered, thick,rounded corners=2pt,
                     minimum height=1.5em, minimum width=5em, inner sep=4pt]
\tikzstyle{contribBlock} = [draw, rectangle, text centered, ultra thick,
					 minimum height=1.5em, 
					 minimum width=5em, inner sep=4pt,
					 dash pattern=on 1pt off 2pt on 4pt off 2pt]
\tikzstyle{typical} = [fill=white!95!black]
\tikzstyle{reddish} = [draw=red,fill=white!95!red]
\tikzstyle{blueish} = [draw=blue,fill=white!95!blue]
\tikzstyle{greenish} = [draw=green!40!gray,fill=white!95!green]
\tikzstyle{orangeish} = [draw=red!40!yellow,fill=white!95!red]

\tikzstyle{longblock} = [draw,rectangle,text centered,thick,rounded corners=2pt,
                     minimum height=1.5em, minimum width=8em, inner sep=4pt]
\tikzstyle{contribLongblock} = [draw, rectangle, text centered, ultra thick,
					 minimum height=1.5em, 
					 minimum width=8em, inner sep=4pt,
					 dash pattern=on 1pt off 2pt on 4pt off 2pt]
\tikzstyle{largeBlock} = [draw, rectangle, thick,
                     minimum height=19em, minimum width=47.5em, inner sep=4pt]
\tikzstyle{smallBlock} = [draw, rectangle, text centered,
                     minimum height=2em, minimum width=4em]
\tikzstyle{dashedBlock} = [draw, dashed, rectangle, ultra thick,
                     minimum height=6.5em, minimum width=25em, inner sep=4pt]
\tikzstyle{thinDashedBlock} = [draw, dashed, rectangle, ultra thick,
                     minimum height=4.25em, minimum width=15em, inner sep=4pt]                     
\tikzstyle{dottedBlock} = [draw, rectangle, text centered, ultra thick,
					 minimum height=1.5em, 
					 minimum width=8em, inner sep=4pt,
					 dash pattern=on 1pt off 2pt on 4pt off 2pt] 
\tikzstyle{notip} = [-, very thick]
\tikzstyle{newtip} = [->, very thick]
\tikzstyle{bidir} = [<->, very thick]
\tikzstyle{newtip_dashed} = [->, very thick, dashed]


\begin{tikzpicture}
    \tikzstyle{every node}=[font=\large]

    \node[block, orangeish, text width=3.5em] (laser) {\centering Laser \\ Scan};
    
    \node[anchor=center,xshift=0.0em,yshift=2.5em] (robot) at (laser.center) {\uline{Robot}};

    \node[block, orangeish, anchor=center, xshift=0.0em,yshift=-8.5em, text width=5.5em] (state) at (laser.center) {\centering State \\ Knowledge};

    \node[anchor=center,xshift=0.0em,yshift=2.5em] (robot) at (state.center) {\uline{Simulation}};

    \node[block, blueish, anchor=center,xshift=8.0em,yshift=0.0em, text width=5.5em] (gap_generation) at (laser) {\centering Gap \\ Generation};

    \node[block, reddish, anchor=center,xshift=8.0em,yshift=6.5em, text width=5.0em] (map) at (laser.center) {Map};

    \node[block, reddish, anchor=center,xshift=8.0em,yshift=0.0em, text width=5.0em] (global_planner) at (map) {\centering Global \\ Planner};

    \node[block, blueish, anchor=center,xshift=8.0em,yshift=0.0em, text width=5.5em] (gap_waypoint_generation) at (gap_generation.center) {\centering Gap Waypoint \\ Generation};

    \node[block, greenish, anchor=center,xshift=0.0em,yshift=-6.0em, text width=5.0em] (obstacle_detection) at (gap_waypoint_generation) {\centering Obstacle \\ Detection};

    \node[block, blueish, anchor=center,xshift=8.0em,yshift=0.0em, text width=5.0em] (waypoint_selection) at (gap_waypoint_generation.center) {\centering Waypoint \\ Selection};

    \node[block, greenish, anchor=center,xshift=8.0em,yshift=0.0em, text width=5.0em] (ga3c_cadrl) at (obstacle_detection.center) {\centering DRL};

    \node[anchor=center,xshift=10.0em,yshift=1.5em] (velocity_command) at (ga3c_cadrl) {Velocity Command};

    \node[anchor=center,xshift=10.0em,yshift=0.0em] (velocity_symbol) at (ga3c_cadrl) {\Large $\mathbf{u_t} = [v_t, \psi_t]$};

    \node[anchor=center, xshift=30em, yshift=3.0em] (top_label) at (global_planner.center)
    {\sc \uline{PLANNING HIERACHY}};

    \node[anchor=center, xshift=0em, yshift=-2.5em] (global_label) at (top_label.center)
    {\sc Global planning};

    \node[anchor=center, xshift=0em, yshift=-6.5em] (local_label) at (global_label.center)
    {\sc Local planning};    

    \node[anchor=center, xshift=0em, yshift=-6.5em] (low_label) at (local_label.center)
    {\sc Low-level control};  

    \draw[newtip] (map.east) -- (global_planner.west);
    \draw[newtip] (global_planner.south) -- (gap_waypoint_generation.north);
    \draw[newtip] ($(laser.east) + (0.125, 2.275)$) -- (map.west);
    
    \draw[newtip] (laser.east) -- (gap_generation.west);
    \draw[notip] ($(laser.east) + (0.125, -2.125)$) -- ($(laser.east) + (0.125, 2.275)$);

    \draw[newtip] (gap_generation.east) -- (gap_waypoint_generation.west);
    \draw[newtip] (gap_waypoint_generation.east) -- (waypoint_selection.west);
    \draw[newtip] (waypoint_selection.south) -- (ga3c_cadrl.north);

    \draw[newtip] ($(laser.east) + (0.125, -2.125)$) -- (obstacle_detection.west);
    \draw[newtip] (obstacle_detection.east) -- (ga3c_cadrl.west);
    \draw[newtip] (ga3c_cadrl.east) -- ($(ga3c_cadrl.east) + (1.0,0.0)$);
    \draw[notip] (state.east) -- ($(state.east) + (7.35, 0.0)$);
    \draw[newtip] ($(state.east) + (7.35, 0.0)$) -- (ga3c_cadrl.south);

    \node[dashedBlock, draw=blue!70!black,anchor=center] (gap_based_planner) at (gap_waypoint_generation.center) {};

    \node[anchor=center,xshift=-8.0em,yshift=2.5em] (gap_based_planner_text) at (gap_based_planner.center) {Waypoint Planner};

    \node[thinDashedBlock, draw=red!70!black,anchor=center] (global_planner_block) at ($(gap_waypoint_generation.center) + (-1.375, 2.25)$) {};

    \node[thinDashedBlock, draw=green!70!black,anchor=center] (velocity_control) at ($(gap_waypoint_generation.center) + (1.375, -2.125)$) {};

\end{tikzpicture}

%% file: RSS_Workshop.bbl
\begin{thebibliography}{30}
\providecommand{\natexlab}[1]{#1}
\providecommand{\url}[1]{\texttt{#1}}
\expandafter\ifx\csname urlstyle\endcsname\relax
  \providecommand{\doi}[1]{doi: #1}\else
  \providecommand{\doi}{doi: \begingroup \urlstyle{rm}\Url}\fi

\bibitem[ELM(1993)]{ELMAN199371}
Learning and development in neural networks: the importance of starting small.
\newblock \emph{Cognition}, 48\penalty0 (1):\penalty0 71--99, 1993.
\newblock ISSN 0010-0277.

\bibitem[Brito et~al.(2021)Brito, Everett, How, and Alonso-Mora]{Brito_Where}
Bruno Brito, Michael Everett, Jonathan~P. How, and Javier Alonso-Mora.
\newblock Where to go next: Learning a subgoal recommendation policy for
  navigation in dynamic environments.
\newblock \emph{IEEE Robotics and Automation Letters}, 6\penalty0 (3):\penalty0
  4616--4623, 2021.

\bibitem[Chen et~al.(2018)Chen, Liu, Kreiss, and Alahi]{chen_crowd_robot_2018}
Changan Chen, Yuejiang Liu, Sven Kreiss, and Alexandre Alahi.
\newblock Crowd-robot interaction: Crowd-aware robot navigation with
  attention-based deep reinforcement learning.
\newblock \emph{CoRR}, abs/1809.08835, 2018.

\bibitem[Chen et~al.(2016)Chen, Liu, Everett, and
  How]{DBLP:journals/corr/ChenLEH16}
Yu~Fan Chen, Miao Liu, Michael Everett, and Jonathan~P. How.
\newblock Decentralized non-communicating multiagent collision avoidance with
  deep reinforcement learning.
\newblock \emph{CoRR}, abs/1609.07845, 2016.

\bibitem[Chen et~al.(2017)Chen, Everett, Liu, and How]{chen_socially_2017}
Yu~Fan Chen, Michael Everett, Miao Liu, and Jonathan~P. How.
\newblock Socially aware motion planning with deep reinforcement learning.
\newblock \emph{CoRR}, abs/1703.08862, 2017.

\bibitem[Chiang et~al.(2018)Chiang, Faust, Fiser, and
  Francis]{chiang_learning_2018}
Hao{-}Tien~Lewis Chiang, Aleksandra Faust, Marek Fiser, and Anthony~G. Francis.
\newblock Learning navigation behaviors end to end.
\newblock \emph{CoRR}, abs/1809.10124, 2018.

\bibitem[Everett et~al.(2018)Everett, Chen, and How]{Everett_Motion}
Michael Everett, Yu~Fan Chen, and Jonathan~P. How.
\newblock Motion planning among dynamic, decision-making agents with deep
  reinforcement learning.
\newblock In \emph{2018 IEEE/RSJ International Conference on Intelligent Robots
  and Systems (IROS)}, pages 3052--3059, 2018.

\bibitem[Everett et~al.(2021)Everett, Chen, and How]{Everett_Collision}
Michael Everett, Yu~Fan Chen, and Jonathan~P. How.
\newblock Collision avoidance in pedestrian-rich environments with deep
  reinforcement learning.
\newblock \emph{IEEE Access}, 9:\penalty0 10357--10377, 2021.

\bibitem[Ferguson et~al.(2008)Ferguson, Darms, Urmson, and Kolski]{4621214}
Dave Ferguson, Michael Darms, Chris Urmson, and Sascha Kolski.
\newblock Detection, prediction, and avoidance of dynamic obstacles in urban
  environments.
\newblock In \emph{2008 IEEE Intelligent Vehicles Symposium}, pages 1149--1154,
  2008.
\newblock \doi{10.1109/IVS.2008.4621214}.

\bibitem[Fiorini and Shiller(1998)]{fiorini1998motion}
Paolo Fiorini and Zvi Shiller.
\newblock Motion planning in dynamic environments using velocity obstacles.
\newblock \emph{The international journal of robotics research}, 17\penalty0
  (7):\penalty0 760--772, 1998.

\bibitem[Ivanovic et~al.(2018)Ivanovic, Harrison, Sharma, Chen, and
  Pavone]{ivanovic_barc_2018}
Boris Ivanovic, James Harrison, Apoorva Sharma, Mo~Chen, and Marco Pavone.
\newblock Barc: Backward reachability curriculum for robotic reinforcement
  learning.
\newblock \emph{CoRR}, abs/1806.06161, 2018.

\bibitem[K{\"a}stner et~al.(2021)K{\"a}stner, Buiyan, Jiao, Le, Zhao, Shen, and
  Lambrecht]{kastner2021arena}
Linh K{\"a}stner, Teham Buiyan, Lei Jiao, Tuan~Anh Le, Xinlin Zhao, Zhengcheng
  Shen, and Jens Lambrecht.
\newblock Arena-rosnav: Towards deployment of deep-reinforcement-learning-based
  obstacle avoidance into conventional autonomous navigation systems.
\newblock In \emph{2021 IEEE/RSJ International Conference on Intelligent Robots
  and Systems (IROS)}, pages 6456--6463. IEEE, 2021.

\bibitem[K{\"a}stner et~al.(2022)K{\"a}stner, Cox, Buiyan, and
  Lambrecht]{9811797}
Linh K{\"a}stner, Johannes Cox, Teham Buiyan, and Jens Lambrecht.
\newblock All-in-one: A drl-based control switch combining state-of-the-art
  navigation planners.
\newblock In \emph{2022 International Conference on Robotics and Automation
  (ICRA)}, pages 2861--2867, 2022.
\newblock \doi{10.1109/ICRA46639.2022.9811797}.

\bibitem[Krantz et~al.(2021)Krantz, Gokaslan, Batra, Lee, and
  Maksymets]{Krantz_Waypoint}
Jacob Krantz, Aaron Gokaslan, Dhruv Batra, Stefan Lee, and Oleksandr Maksymets.
\newblock Waypoint models for instruction-guided navigation in continuous
  environments.
\newblock In \emph{Proceedings of the IEEE/CVF International Conference on
  Computer Vision (ICCV)}, pages 15162--15171, October 2021.

\bibitem[Kästner et~al.(2023)Kästner, Meusel, Bhuiyan, and
  Lambrecht]{kästner2023holistic}
Linh Kästner, Marvin Meusel, Teham Bhuiyan, and Jens Lambrecht.
\newblock Holistic deep-reinforcement-learning-based training of autonomous
  navigation systems, 2023.

\bibitem[Long et~al.(2018)Long, Fanl, Liao, Liu, Zhang, and
  Pan]{long_towards_2018}
Pinxin Long, Tingxiang Fanl, Xinyi Liao, Wenxi Liu, Hao Zhang, and Jia Pan.
\newblock Towards optimally decentralized multi-robot collision avoidance via
  deep reinforcement learning.
\newblock In \emph{2018 IEEE International Conference on Robotics and
  Automation (ICRA)}, page 6252–6259. IEEE Press, 2018.

\bibitem[Long et~al.(2020)Long, Zhou, Gupta, Fang, Wu, and
  Wang]{long_evolutionary_2020}
Qian Long, Zihan Zhou, Abhinav Gupta, Fei Fang, Yi~Wu, and Xiaolong Wang.
\newblock Evolutionary population curriculum for scaling multi-agent
  reinforcement learning.
\newblock \emph{CoRR}, abs/2003.10423, 2020.

\bibitem[Marder-Eppstein(2020)]{ros_move_base}
Eitan Marder-Eppstein.
\newblock Ros move\_base package.
\newblock 2020.

\bibitem[Missura and Bennewitz(2019)]{maren_DWA}
Marcell Missura and Maren Bennewitz.
\newblock Predictive collision avoidance for the dynamic window approach.
\newblock pages 8620--8626, 05 2019.
\newblock \doi{10.1109/ICRA.2019.8794386}.

\bibitem[Quigley et~al.(2009)Quigley, Gerkey, Conley, Faust, Foote, Leibs,
  Berger, Wheeler, and Ng]{quigley_ros_2009}
Morgan Quigley, Brian Gerkey, Ken Conley, Josh Faust, Tully Foote, Jeremy
  Leibs, Eric Berger, Rob Wheeler, and Andrew Ng.
\newblock {ROS}: an open-source {Robot} {Operating} {System}.
\newblock In \emph{Proc. of the {IEEE} {Intl}. {Conf}. on {Robotics} and
  {Automation} ({ICRA}) {Workshop} on {Open} {Source} {Robotics}}, Kobe, Japan,
  May 2009.

\bibitem[Rajeswar et~al.(2017)Rajeswar, Subramanian, Dutil, Pal, and
  Courville]{curr_nlp}
Sai Rajeswar, Sandeep Subramanian, Francis Dutil, Christopher~Joseph Pal, and
  Aaron~C. Courville.
\newblock Adversarial generation of natural language.
\newblock \emph{CoRR}, abs/1705.10929, 2017.

\bibitem[Semnani et~al.(2020)Semnani, Liu, Everett, de~Ruiter, and
  How]{Semnani_Multi_agent}
Samaneh~Hosseini Semnani, Hugh Liu, Michael Everett, Anton de~Ruiter, and
  Jonathan~P. How.
\newblock Multi-agent motion planning for dense and dynamic environments via
  deep reinforcement learning.
\newblock \emph{IEEE Robotics and Automation Letters}, 5\penalty0 (2):\penalty0
  3221--3226, 2020.

\bibitem[Soviany et~al.(2021)Soviany, Ionescu, Rota, and Sebe]{curriculum}
Petru Soviany, Radu~Tudor Ionescu, Paolo Rota, and Nicu Sebe.
\newblock Curriculum learning: {A} survey.
\newblock \emph{CoRR}, abs/2101.10382, 2021.

\bibitem[Tai et~al.(2018)Tai, Zhang, Liu, and Burgard]{tai_socially_2018}
Lei Tai, Jingwei Zhang, Ming Liu, and Wolfram Burgard.
\newblock Socially compliant navigation through raw depth inputs with
  generative adversarial imitation learning.
\newblock In \emph{2018 IEEE International Conference on Robotics and
  Automation (ICRA)}, pages 1111--1117, 2018.
\newblock \doi{10.1109/ICRA.2018.8460968}.

\bibitem[Tsardoulias and Zalidis(2014)]{tsardoulias_stdr_simulator_2014}
A.~T.~M. Tsardoulias and C.~Zalidis.
\newblock stdr\_simulator - ros\_wiki, 2014.
\newblock URL \url{http://wiki.ros.org/stdr\_simulator}.

\bibitem[van~den Berg et~al.(2011)van~den Berg, Guy, Lin, and
  Manocha]{vandenburg_recipriocal_2011}
Jur van~den Berg, Stephen~J. Guy, Ming Lin, and Dinesh Manocha.
\newblock Reciprocal n-body collision avoidance.
\newblock In C{\'e}dric Pradalier, Roland Siegwart, and Gerhard Hirzinger,
  editors, \emph{Robotics Research}, pages 3--19, Berlin, Heidelberg, 2011.
  Springer Berlin Heidelberg.

\bibitem[Wang et~al.(2019{\natexlab{a}})Wang, Lehman, Clune, and
  Stanley]{wang_poet_2019}
Rui Wang, Joel Lehman, Jeff Clune, and Kenneth~O. Stanley.
\newblock Paired open-ended trailblazer {(POET):} endlessly generating
  increasingly complex and diverse learning environments and their solutions.
\newblock \emph{CoRR}, abs/1901.01753, 2019{\natexlab{a}}.

\bibitem[Wang et~al.(2019{\natexlab{b}})Wang, Gan, Yang, Wu, and Yan]{9008373}
Yiru Wang, Weihao Gan, Jie Yang, Wei Wu, and Junjie Yan.
\newblock Dynamic curriculum learning for imbalanced data classification.
\newblock In \emph{2019 IEEE/CVF International Conference on Computer Vision
  (ICCV)}, pages 5016--5025, 2019{\natexlab{b}}.
\newblock \doi{10.1109/ICCV.2019.00512}.

\bibitem[Xu et~al.(2021)Xu, Feng, and Vela]{Xu_Potential}
Ruoyang Xu, Shiyu Feng, and Patricio~A. Vela.
\newblock Potential gap: A gap-informed reactive policy for safe hierarchical
  navigation.
\newblock \emph{IEEE Robotics and Automation Letters}, 6\penalty0 (4):\penalty0
  8325--8332, 2021.

\bibitem[Zhang et~al.(2020)Zhang, Abbeel, and Pinto]{curr_value}
Yunzhi Zhang, Pieter Abbeel, and Lerrel Pinto.
\newblock Automatic curriculum learning through value disagreement.
\newblock \emph{CoRR}, abs/2006.09641, 2020.

\end{thebibliography}
